\documentclass[11pt,a4paper]{article}
\usepackage[hyperref]{acl2017}
\usepackage{times}
\usepackage{latexsym}
\usepackage{url}
\usepackage{subcaption}
\usepackage{graphicx}
\usepackage{placeins}

\DeclareCaptionFont{tiny}{\tiny}

\usepackage{amsmath,amsthm,amssymb,amsfonts,mathtools}

\usepackage{tabularx}
\usepackage{multirow}
\usepackage{wrapfig}

\usepackage[justification=justified,singlelinecheck=false]{caption}

\usepackage{algorithm}
\usepackage{algpseudocode}

\usepackage{tikz-dependency}
\usepackage{tikz}
\usetikzlibrary{positioning}
\definecolor {processblue}{cmyk}{0.96,0,0,0}

\newcommand{\textoverline}[1]{$\overline{\mbox{#1}}$}

\newcommand{\APT}{%
	\textsc{Apt}}

\aclfinalcopy 
\def\aclpaperid{3206} 

\title{Improving Semantic Composition with Offset Inference}

\author{
  Thomas Kober, Julie Weeds, Jeremy Reffin \and David Weir\\
 TAG laboratory, Department of Informatics, University of Sussex\\
 Brighton, BN1 9RH, UK\\
  {\tt \{t.kober, j.e.weeds, j.p.reffin, d.j.weir\}@sussex.ac.uk}
}

\date{}

\begin{document}
\maketitle
\begin{abstract}
Count-based distributional semantic models suffer from sparsity due to unobserved but plausible co-occurrences in any text collection. This problem is amplified for models like Anchored Packed Trees ({\APT}s), that take the grammatical type of a co-occurrence into account. We therefore introduce a novel form of distributional inference that exploits the rich type structure in {\APT}s and infers missing data by the same mechanism that is used for semantic composition.
\end{abstract}

\section{Introduction}
Anchored Packed Trees ({\APT}s) is a recently proposed approach to distributional semantics that takes distributional composition to be a process of lexeme contextualisation~\citep{Weir_2016}. A lexeme's meaning, characterised as  knowledge concerning co-occurrences involving that lexeme, is represented with a higher-order dependency-typed structure (the {\APT})  where paths associated with higher-order dependencies connect vertices associated with weighted lexeme multisets. The central innovation in the compositional theory is that the {\APT}'s type structure enables the precise alignment of the semantic representation of each of the lexemes being composed. Like other count-based distributional spaces, however, it is prone to considerable data sparsity, caused by not observing all plausible co-occurrences in the given data. Recently, \citet{Kober_2016} introduced a simple unsupervised algorithm to infer missing co-occurrence information by leveraging the distributional neighbourhood and ease the sparsity effect in count-based models. 

In this paper, we generalise distributional inference (DI) in {\APT}s and show how precisely the same mechanism that was introduced to support distributional composition, namely ``offsetting" {\APT}~representations, gives rise to a novel form of distributional inference, allowing us to infer co-occurrences from neighbours of these representations. For example, by transforming a representation of \emph{white} to a 
representation of ``things that can be white", inference of unobserved, but plausible, co-occurrences can be based 
on finding near neighbours (which will be nouns) of the ``things that can be white" structure. This furthermore exposes an interesting connection between distributional inference and distributional composition. Our method is unsupervised and maintains the intrinsic interpretability of {\APT}s\footnote{We release our code and data at \url{https://github.com/tttthomasssss/acl2017}}.
\label{introduction}

\section{Offset Representations}
The basis of how composition is modelled in the {\APT}~framework is the way that the co-occurrences are structured. In characterising the distributional semantics of some lexeme $w$, rather than just recording a co-occurrence between $w$ and $w^{\prime}$ within some context window, we follow~\citet{Pado_2007} and record the dependency path from $w$ to $w^{\prime}$. This syntagmatic structure makes it possible to appropriately offset the semantic representations of each of the lexemes being composed in some phrase. For example many nouns will have distributional features starting with the type \texttt{amod}, which cannot be observed for adjectives or verbs. Thus, when composing the adjective \emph{white} with the noun \emph{clothes}, the feature spaces of the two lexemes need to be aligned first. This can be achieved by offsetting one of the constituents, which we will explain in more detail in this section. 

We will make use of the following notation throughout this work. A typed distributional feature consists of a path and a lexeme such as in \texttt{amod:}\emph{white}. Inverse paths are denoted by a horizontal bar above the dependency relation such as in \texttt{\textoverline{dobj}:}\emph{prefer} and higher-order paths are separated by a dot such as in \texttt{\textoverline{amod}.\textoverline{compound}:}\emph{dress}. 

\begin{figure*}[!htb]
\includegraphics[width=\textwidth]{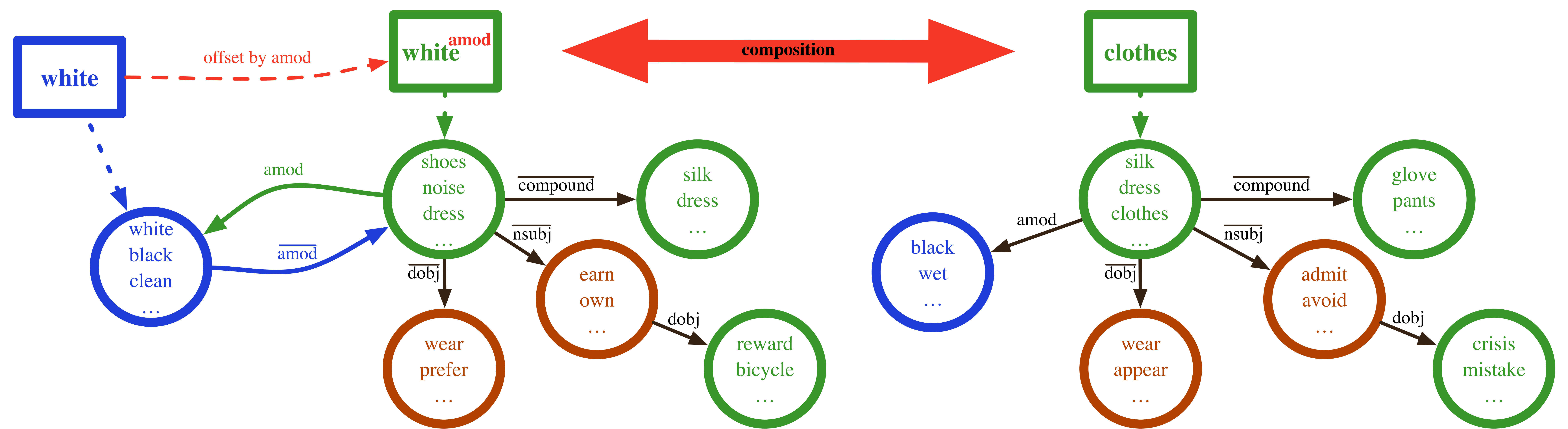}
\captionsetup{font=small}
\caption{Structured distributional \APT~space. Different colours reflect different parts of speech. Boxes denote the current anchor of the \APT, circles represent nodes in the \APT~space, holding lexemes, and edges represent their relationship within the space.}
\label{apt_space}
\end{figure*}

Offset representations are the central component in the composition process in the {\APT}~framework. Figure~\ref{apt_space} shows the \APT~representations for the adjective \emph{white} (left) and the \APT~for the noun \emph{clothes} (right), as might have been observed in a text collection. Each node holds a multiset of lexemes and the anchor of an {\APT}~reflects the current perspective of a lexeme at the given node. An offset representation can be created by shifting the anchor along a given path. For example the lexeme \emph{white} is at the same node as other adjectives such as \emph{black} and \emph{clean}, whereas nouns such as \emph{shoes} or \emph{noise} are typically reached via the \texttt{\textoverline{amod}} edge. 

Offsetting in {\APT}s only involves a change in the anchor, the underlying structure remains unchanged. By offsetting the lexeme \emph{white} by \texttt{amod} the anchor is shifted along the \texttt{\textoverline{amod}} edge, which results in creating a noun view for the adjective \emph{white}. We denote the offset view of a lexeme for a given path by superscripting the offset path, for example the \texttt{amod} offset of the adjective \emph{white} is denoted as $\emph{white}^\texttt{amod}$. The offsetting procedure changes the starting points of the paths as visible in Figure~\ref{apt_space} between the anchors for \emph{white} and $\emph{white}^\texttt{amod}$, since paths always begin at the anchor. The red dashed line in Figure~\ref{apt_space} reflects that anchor shift. The lexeme $\emph{white}^\texttt{amod}$ represents a prototypical ``white thing", that is, a noun that has been modified by the adjective \emph{white}. We note that all edges in the {\APT}~space are bi-directional as exemplified in the coloured \texttt{amod} and \texttt{\textoverline{amod}} edges in the \APT~for \emph{white}, however for brevity we only show uni-directional edges in Figure~\ref{apt_space}.

By considering the \APT~representations for the lexemes \emph{white} and \emph{clothes} in Figure~\ref{apt_space}, it becomes apparent that lexemes with different parts of speech are located in different areas of the semantic space. If we want to compose the adjective-noun phrase \emph{white clothes}, we need to offset one of the two constituents to align the feature spaces in order to leverage their distributional commonalities. This can be achieved by either creating a noun offset view of \emph{white}, by shifting the anchor along the \texttt{\textoverline{amod}} edge, or by creating an adjective offset representation of \emph{clothes} by shifting its anchor along \texttt{amod}. In this work we follow~\citet{Weir_2016} and always offset the dependent in a given relation. Table~\ref{apt_feature_space} shows a subset of the features of Figure~\ref{apt_space} as would be represented in a vectorised \APT. Vectorising the whole~\APT~lexicon results in a very high-dimensional and sparse typed distributional space. The features for $\emph{white}^\texttt{amod}$ (middle column) highlight the change in feature space caused by offsetting the adjective \emph{white}. The features of the offset view $\emph{white}^\texttt{amod}$, are now aligned with the noun \emph{clothes} such that the two can be composed. Composition can be performed by either selecting the \emph{union} or \emph{intersection} of the aligned features. 
\begin{table}[!htb]
\centering
\small
\resizebox{\columnwidth}{!}{
\begin{tabular}{ l | l || l }
\multicolumn{1}{c|}{\textbf{\emph{white}}} & \multicolumn{1}{c||}{\textbf{$\emph{white}^\texttt{amod}$}} & \multicolumn{1}{c}{\textbf{\emph{clothes}}}\\ \hline
\texttt{:}\emph{clean} & \texttt{amod:}\emph{clean} & \texttt{amod:}\emph{wet} \\
\texttt{\textoverline{amod}:}\emph{shoes} & \texttt{:}\emph{shoes} & \texttt{:}\emph{dress} \\
\texttt{\textoverline{amod}.\textoverline{dobj}:}\emph{wear} & \texttt{\textoverline{dobj}:}\emph{wear} & \texttt{\textoverline{dobj}:}\emph{wear} \\
\texttt{\textoverline{amod}.\textoverline{nsubj}.}\emph{earn} & \texttt{\textoverline{nsubj}:}\emph{earn} & \texttt{\textoverline{nsubj}:}\emph{admit} \\
\end{tabular}}
\captionsetup{font=small}
\caption{Sample of vectorised features for the {\APT}s shown in Figure~\ref{apt_space}. Offsetting \emph{white} by \texttt{amod} creates an offset view, $\emph{white}^\texttt{amod}$, representing a noun, and has the consequence of aligning the feature space with \emph{clothes}.}
\label{apt_feature_space}
\end{table}

\subsection{Qualitative Analysis of Offset Representations}

Any offset view of a lexeme is behaviourally identical to a ``normal" lexeme. It has an associated part of speech, a distributional representation which locates it in semantic space, and we can find neighbours for it in the same way that we find neighbours for any other lexeme. In this way, a single {\APT} data structure is able to provide many different views of any given lexeme. These views reflect the different ways in which the lexeme is used. For example $\emph{law}^\texttt{\textoverline{nsubj}}$ is the \texttt{\textoverline{nsubj}} offset representation of the noun \emph{law}. This lexeme is a verb and represents an action carried out by the \emph{law}. This contrasts with $\emph{law}^\texttt{\textoverline{dobj}}$, which is the \texttt{\textoverline{dobj}} offset representation of the noun \emph{law}.  It is also a verb, however represents actions done to the \emph{law}. Table~\ref{offset_nearest_neighbours} lists the $10$ nearest neighbours for a number of lexemes, offset by \texttt{amod}, \texttt{\textoverline{dobj}} and \texttt{\textoverline{nsubj}} respectively.

For example, the neighbourhood of the lexeme \emph{ancient} in Table~\ref{offset_nearest_neighbours} shows that the offset view for $\emph{ancient}^\texttt{amod}$ is a prototypical representation of an ``ancient thing", with neighbours easily associated with the property \emph{ancient}. Furthermore, Table~\ref{offset_nearest_neighbours} illustrates that nearest neighbours of offset views are often other offset representations. This means that for example actions carried out by a \emph{mother} tend to be similar to actions carried out by a \emph{father} or a \emph{parent}. 
\begin{table*}[!htb]
\centering
\small
\resizebox{\textwidth}{!}{
\begin{tabular}{ l | l }
\textbf{Offset Representation} & \textbf{Nearest Neighbours}\\ \hline
$\emph{ancient}^\texttt{amod}$ & civilzation, mythology, tradition, ruin, monument, trackway, tomb, antiquity, folklore, deity\\
$\emph{red}^\texttt{amod}$ & $\text{blue}^\texttt{amod}$, $\text{black}^\texttt{amod}$, $\text{green}^\texttt{amod}$, $\text{dark}^\texttt{amod}$, onion, pepper, red, tomato, carrot, garlic \\ 
$\emph{economic}^\texttt{amod}$ & $\text{political}^\texttt{amod}$, $\text{societal}^\texttt{amod}$, cohabiting, economy, growth, cohabitant, globalisation, competitiveness, \\& globalization, prosperity \\ \hline
$\emph{government}^{\texttt{\textoverline{dobj}}}$ & overthrow, $\text{party}^{\texttt{\textoverline{dobj}}}$, $\text{authority}^{\texttt{\textoverline{dobj}}}$, $\text{leader}^{\texttt{\textoverline{dobj}}}$, $\text{capital}^{\texttt{\textoverline{dobj}}}$, $\text{force}^{\texttt{\textoverline{dobj}}}$, $\text{state}^{\texttt{\textoverline{dobj}}}$, $\text{official}^{\texttt{\textoverline{dobj}}}$, $\text{minister}^{\texttt{\textoverline{dobj}}}$, oust \\
$\emph{problem}^\texttt{\textoverline{dobj}}$ & $\text{difficulty}^\texttt{\textoverline{dobj}}$, solve, coded, $\text{issue}^\texttt{\textoverline{dobj}}$, $\text{injury}^\texttt{\textoverline{dobj}}$, overcome,  $\text{question}^\texttt{\textoverline{dobj}}$,  think,  $\text{loss}^\texttt{\textoverline{dobj}}$, relieve \\
$\emph{law}^{\texttt{\textoverline{dobj}}}$ & violate, $\text{rule}^\texttt{\textoverline{dobj}}$, enact, repeal, $\text{principle}^{\texttt{\textoverline{dobj}}}$, unmake, enforce, $\text{policy}^{\texttt{\textoverline{dobj}}}$, obey, flout \\ \hline
$\emph{researcher}^\texttt{\textoverline{nsubj}}$  & $\text{physician}^\texttt{\textoverline{nsubj}}$, $\text{writer}^\texttt{\textoverline{nsubj}}$, theorize, thwart, theorise, hypothesize, surmise, $\text{student}^\texttt{\textoverline{nsubj}}$, $\text{worker}^\texttt{\textoverline{nsubj}}$, apprehend \\
$\emph{mother}^\texttt{\textoverline{nsubj}}$ & $\text{wife}^\texttt{\textoverline{nsubj}}$, $\text{father}^\texttt{\textoverline{nsubj}}$, $\text{parent}^\texttt{\textoverline{nsubj}}$, $\text{woman}^\texttt{\textoverline{nsubj}}$, re-married, remarry, $\text{girl}^\texttt{\textoverline{nsubj}}$,breastfeed, $\text{family}^\texttt{\textoverline{nsubj}}$, disown \\
$\emph{law}^{\texttt{\textoverline{nsubj}}}$ & $\text{rule}^{\texttt{\textoverline{nsubj}}}$, $\text{principle}^{\texttt{\textoverline{nsubj}}}$, $\text{policy}^{\texttt{\textoverline{nsubj}}}$, criminalize, $\text{case}^{\texttt{\textoverline{nsubj}}}$, $\text{contract}^{\texttt{\textoverline{nsubj}}}$, prohibit, proscribe, enjoin, $\text{charge}^\texttt{\textoverline{nsubj}}$
\end{tabular}}
\captionsetup{font=small}
\caption{List of the $10$ nearest neighbours of \texttt{amod}, \texttt{\textoverline{dobj}} and \texttt{\textoverline{nsubj}} offset representations.}
\label{offset_nearest_neighbours}
\end{table*}

\subsection{Offset Inference}
Our approach generalises the unsupervised algorithm proposed by~\citet{Kober_2016}, henceforth ``standard DI", as a method for inferring missing knowledge into an {\APT}~representation. Rather than simply inferring potentially plausible, but unobserved co-occurrences from near distributional neighbours, inferences can be made involving offset {\APT}s. For example, the adjective \emph{white} can be offset so that it represents a noun --- a prototypical ``white thing". This allows inferring plausible co-occurrences from other ``things that can be white", such as \emph{shoes} or \emph{shirts}. Our algorithm therefore reflects the contextualised use of a word. This has the advantage of being able to make flexible and fine grained distinctions in the inference process. For example if the noun \emph{law} is used as a subject, our algorithm allows inferring plausible co-occurrences from ``other actions carried out by the law". This contrasts the use of \emph{law} as an object, where offset inference is able to find co-occurrences on the basis of ``other actions done to the law". This is a crucial advantage over the method of~\citet{Kober_2016} which only supports inference on uncontextualised lexemes.

A sketch of how offset inference for a lexeme $w$ works is shown in Algorithm~\ref{alg:offset_inference}. Our algorithm requires a distributional model $M$, an {\APT} representation for the lexeme $w$ for which to perform offset inference, a dependency path $p$, describing the offset for $w$, and the number of neighbours $k$. The offset representation of $w^{\prime}$ is then enriched with the information from its distributional neighbours by some merge function. We note that if the offset path $p$ is the empty path, we would recover the algorithm presented by~\citet{Kober_2016}. Our algorithm is unsupervised, and agnostic to the input distributional model and the neighbour retrieval function. 
\begin{algorithm}
	\caption{Offset Inference}
	\label{alg:offset_inference}
	\begin{algorithmic}[1]
		\Procedure{offset\_inference}{$M$, $w$, $p$, $k$}
			\State $w^{\prime} \gets \text{offset}(w, p)$
			\ForAll{$n$ in $\text{neighbours}(M, w^{\prime}, k)$}
				\State $w^{\prime\prime} \gets \text{merge}(w^{\prime\prime}, n)$
			\EndFor
			\State \textbf{return} $w^{\prime\prime}$
		\EndProcedure
	\end{algorithmic}
\end{algorithm}

\subsection*{Connection to Distributional Composition}
An interesting observation is the similarity between distributional inference and distributional composition, as both operations are realised by the same mechanism --- an offset followed by inferring plausible co-occurrence counts for a single lexeme in the case of distributional inference, or for a phrase in the case of composition. The merging of co-occurrence dimensions for distributional inference can also be any of the operations commonly used for distributional composition such as pointwise minimum, maximum, addition or multiplication. 

This relation creates an interesting dynamic between distributional inference and composition when used in a complementary manner as in this work. The former can be used as a process of \emph{co-occurrence embellishment} which is adding missing information, however with the risk of introducing some noise. The latter on the other hand can be used as a process of \emph{co-occurrence filtering}, that is leveraging the enriched representations, while also sieving out the previously introduced noise.
\label{offset_inference}

\section{Experiments}
For our experiments we re-implemented the standard DI method of~\citet{Kober_2016} for a direct comparison. We built an order 2 \APT~space on the basis of the concatenation of ukWaC, Wackypedia and the BNC~\citep{Baroni_2009}, pre-parsed with the Malt parser~\citep{Nivre_2006b}. We PPMI transformed the raw co-occurrence counts prior to composition, using a negative SPPMI shift of $\log 5$~\citep{Levy_2014c}. We also experimented with composing normalised counts and applying the PPMI transformation after composition as done by~\citet{Weeds_2017}, however found composing PPMI scores to work better for this task. 

We evaluate our offset inference algorithm on two popular short phrase composition benchmarks by~\citet{Mitchell_2008} and~\citet{Mitchell_2010}, henceforth ML08 and ML10 respectively. The ML08 dataset consists of 120 distinct verb-object (VO) pairs and the ML10 dataset contains 108 adjective-noun (AN), 108 noun-noun (NN) and 108 verb-object pairs. The goal is to compare a model's similarity estimates to human provided judgements. For both tasks, each phrase pair has been rated by multiple human annotators on a scale between 1 and 7, where 7 indicates maximum similarity. Comparison with human judgements is achieved by calculating Spearman's $\rho$ between the model's similarity estimates and the scores of each human annotator individually. We performed composition by \emph{intersection} and tuned the number of neighbours by a grid search over \{0, 10, 30, 50, 100, 500, 1000\} on the ML10 development set, selecting 10 neighbours for NNs, 100 for ANs and 50 for VOs for both DI algorithms. We calculate statistical significance using the method of~\citet{Steiger_1980}.

\subsubsection*{Effect of the number of neighbours}
\begin{figure}[!htb]
\includegraphics[width=\columnwidth]{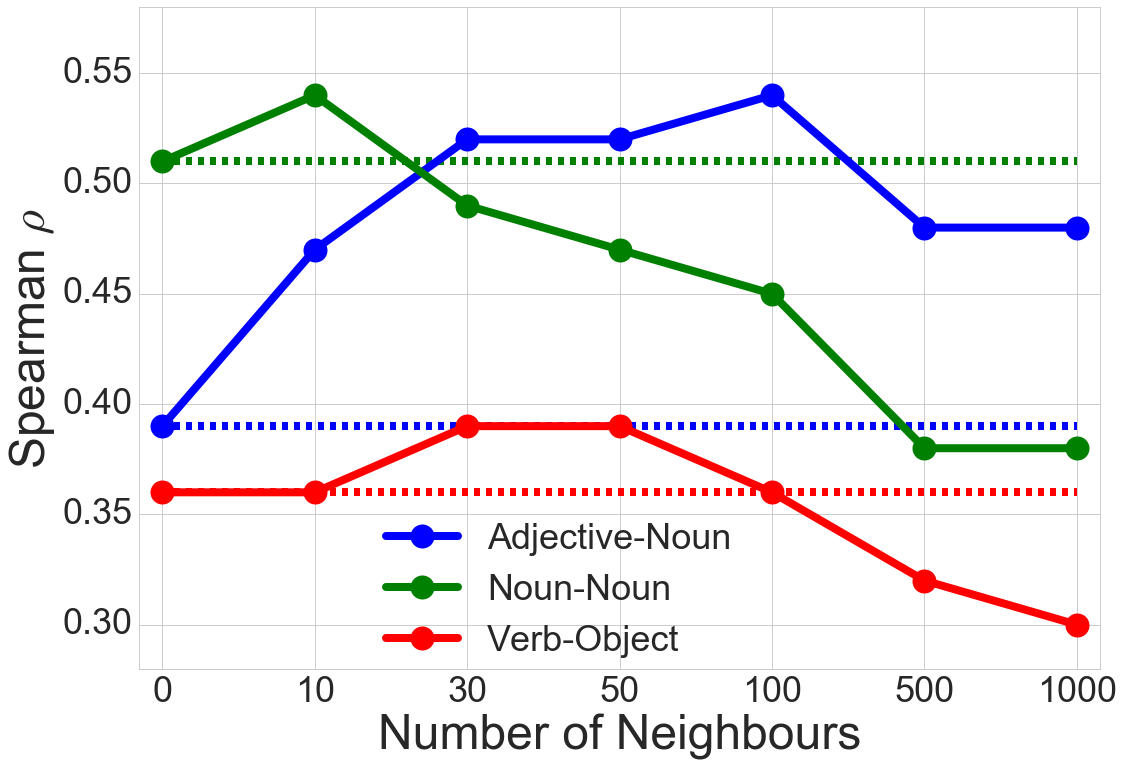}
\captionsetup{font=small}
\caption{Effect of the number of neighbours on the ML10 development set.}
\label{effect_of_neighbours}
\end{figure}

Figure~\ref{effect_of_neighbours} shows the effect of the number of neighbours for AN, NN and VO phrases, using offset inference, on the ML10 development set. Interestingly, NN compounds exhibit an early saturation effect, while VOs and ANs require more neighbours for optimal performance. One explanation for the observed behaviour is that up to some threshold, the neighbours being added contribute actually missing co-occurrence events, whereas past that threshold distributional inference degrades to just generic smoothing that is simply compensating for sparsity, but overwhelming the representations with non-plausible co-occurrence information. A similar effect has also been observed by~\citet{Erk_2010b} in an exemplar-based model.

\subsubsection*{Results}

Table~\ref{composition_distributional_inference} shows that both forms of distributional inference significantly outperform a baseline without DI. On average, offset inference outperforms the method of \citet{Kober_2016} by a statistically significant margin on both datasets.
\begin{table}[!htb]
\centering
\small
\resizebox{\columnwidth}{!}{
\begin{tabular}{ l | c | c | c | c | c}
& \multicolumn{4}{c|}{\textbf{ML10}} & \textbf{ML08} \\ \hline
\textbf{\APT~configuration} & \textbf{AN} & \textbf{NN} & \textbf{VO} & \textbf{Avg} & \textbf{VO} \\ \hline
None & $0.35$ & $0.50$ & $0.39$ & $0.41$ & $0.22$\\
Standard DI & $0.48^{\ddagger}$ & $0.51$ & $0.43^{\ddagger}$ & $0.47^{\ddagger}$ & $0.29^{\ddagger}$ \\
Offset Inference & $\textbf{0.49}^{\ddagger}$ & $\textbf{0.52}$ & $\textbf{0.44}^{\ddagger}$ & $\textbf{0.48}^{*\ddagger}$ & $\textbf{0.31}^{\dagger\ddagger}$ \\
\end{tabular}}
\captionsetup{font=small}
\caption{Comparison of DI algorithms. $\ddagger$ denotes statistical significance at $p < 0.01$ in comparison to the method without DI, * denotes statistical significance at $p < 0.01$ in comparison to standard DI and $\dagger$ denotes statistical significance at $p < 0.05$ in comparison to standard DI.}
\label{composition_distributional_inference}
\end{table}

Table~\ref{composition_distributional_inference_overview} shows that offset inference substantially outperforms comparable sparse models by~\citet{Dinu_2013} on ML08, achieving a new state-of-the-art, and matches the performance of the state-of-the-art neural network model of \citet{Hashimoto_2014} on ML10, while being fully interpretable.
\begin{table}[!htb]
\centering
\small
\resizebox{\columnwidth}{!}{
\begin{tabular}{ l | c | c }
\textbf{Model} & \textbf{ML10 - Average} & \textbf{ML08} \\ \hline
Our work & $\textbf{0.48}$ & $\textbf{0.31}$\\
\citet{Blacoe_2012} & $0.44$ & -\\
\citet{Hashimoto_2014} & $\textbf{0.48}$ & - \\
\citet{Weir_2016} & $0.43$ & $0.26$ \\
\citet{Dinu_2013} & - & $0.23-0.26$ \\
\citet{Erk_2008} & - & $0.27$
\end{tabular}}
\captionsetup{font=small}
\caption{Comparison with existing methods.}
\label{composition_distributional_inference_overview}
\end{table}

\label{experiments}

\section{Related Work}
Distributional inference has its roots in the work of~\citet{Dagan_1993,Dagan_1994}, who aim to find probability estimates for unseen words in bigrams, and~\citet{Schutze_1992,Schutze_1998} who leverages the distributional neighbourhood through clustering of contexts for word-sense discrimination. Recently~\citet{Kober_2016} revitalised the idea for compositional distributional semantic models.

Composition with distributional semantic models has become a popular research area in recent years. Simple, yet  competitive methods, are based on pointwise vector addition or multiplication~\citep{Mitchell_2008,Mitchell_2010}.  However, these approaches neglect the structure of the text defining composition as a commutative operation.

A number of approaches proposed in the literature  attempt to overcome this shortcoming by introducing weighted additive variants~\citep{Guevara_2010,Guevara_2011,Zanzotto_2010}. Another popular strand of work models semantic composition on the basis of ideas arising in formal semantics. Composition in such models is usually implemented as operations on higher-order tensors~\citep{Baroni_2010,Baroni_2014b,Coecke_2011,Grefenstette_2011,Grefenstette_2011b,Grefenstette_2013,Kartsaklis_2014,Paperno_2014,Tian_2016,VanDeCruys_2013}. 
Another widespread approach to semantic composition is to use neural networks~\citep{Bowman_2016,Hashimoto_2014,Hill_2016,Mou_2015,Socher_2012,Socher_2014,Wieting_2015,Yu_2015}, or convolutional tree kernels~\citep{Croce_2011,Zanzotto_2011,Annesi_2014} as composition functions.

The above approaches are applied to untyped distributional vector space models where untyped models contrast with typed models~\citep{Baroni_2010b} in terms of whether structural information is encoded in the representation as in the models of~\citet{Erk_2008,Gamallo_2017,Levy_2014,Pado_2007,Thater_2010,Thater_2011,Weeds_2014}.

The perhaps most popular approach in the literature to evaluating compositional distributional semantic models is to compare human word and phrase similarity judgements with similarity estimates of composed meaning representations, under the assumption that better distributional representations will perform better at these tasks~\citep{Blacoe_2012,Dinu_2013,Erk_2008,Hashimoto_2014,Hermann_2013,Kiela_2014,Turney_2012}.
\label{related_work}

\section{Conclusion}
In this paper we have introduced a novel form of distributional inference that generalises the method introduced by~\citet{Kober_2016}. We have shown its effectiveness for semantic composition on two benchmark phrase similarity tasks where we achieved state-of-the-art performance while retaining the interpretability of our model. We have furthermore highlighted an interesting connection between distributional inference and distributional composition.

In future work we aim to apply our novel method to improve modelling selectional preferences, lexical inference, and scale up to longer phrases and full sentences.
\label{conclusion}

\clearpage
\bibliographystyle{acl_natbib}
\bibliography{common}

\end{document}